\documentclass[letterpaper, 10 pt, journal]{ieeeconf} 

\title{Graphical Games for UAV Swarm Control Under Time-Varying Communication Networks}
\author{Paper ID: }
\date{April 2021}
\IEEEoverridecommandlockouts                              

\usepackage[utf8]{inputenc}
\usepackage{amsmath}
\usepackage[Symbol]{upgreek}
\usepackage{graphics}
\usepackage[tight]{subfigure}
\usepackage{hyperref}
\usepackage{bm}
\usepackage[]{footmisc}
\usepackage[ruled,vlined,linesnumbered]{algorithm2e}
\usepackage{comment}
\usepackage{todonotes}
\usepackage{amssymb}
\usepackage{amsfonts}

\usepackage{amsthm}
\usepackage{blindtext}
\usepackage{cite}

\usepackage{scalerel}
\usepackage{tikz}
\usetikzlibrary{svg.path}

\definecolor{orcidlogocol}{HTML}{A6CE39}
\tikzset{
  orcidlogo/.pic={
    \fill[orcidlogocol] svg{M256,128c0,70.7-57.3,128-128,128C57.3,256,0,198.7,0,128C0,57.3,57.3,0,128,0C198.7,0,256,57.3,256,128z};
    \fill[white] svg{M86.3,186.2H70.9V79.1h15.4v48.4V186.2z}
                 svg{M108.9,79.1h41.6c39.6,0,57,28.3,57,53.6c0,27.5-21.5,53.6-56.8,53.6h-41.8V79.1z M124.3,172.4h24.5c34.9,0,42.9-26.5,42.9-39.7c0-21.5-13.7-39.7-43.7-39.7h-23.7V172.4z}
                 svg{M88.7,56.8c0,5.5-4.5,10.1-10.1,10.1c-5.6,0-10.1-4.6-10.1-10.1c0-5.6,4.5-10.1,10.1-10.1C84.2,46.7,88.7,51.3,88.7,56.8z};
  }
}

\newcommand\orcidicon[1]{\href{https://orcid.org/#1}{\mbox{\scalerel*{
\begin{tikzpicture}[yscale=-1,transform shape]
\pic{orcidlogo};
\end{tikzpicture}
}{|}}}}
\usepackage{hyperref}

\usepackage{eso-pic}
\newcommand\AtPageUpperMyright[1]{\AtPageUpperLeft{
 \put(\LenToUnit{0.5\paperwidth},\LenToUnit{-1cm}){
     \parbox{0.5\textwidth}{\raggedleft\fontsize{10}{11}\selectfont #1} }
 }}
\newcommand{\conf}[1]{
\AddToShipoutPictureBG*{
\AtPageUpperMyright{#1}
}
}

\graphicspath{{./figs/}}

\title{\LARGE \bf
Graphical Games for UAV Swarm Control Under \\Time-Varying Communication Networks
}
\author{Malintha~Fernando, Ransalu~Senanayake, Ariful Azad, Martin~Swany
\thanks{Malintha Fernando, Ariful Azad and Martin Swany are with the Luddy School of Informatics, Computing, and Engineering  at Indiana University, Bloomngton, IN, 47401, USA. E-mail:{\tt\small \{ccfernan, azad, swany\} @iu.edu, ransalu@stanford.edu} }
\thanks{Ransalu Senanayake is with Stanford University, CA, 94305, USA. E-mail:{\tt\small ransalu@stanford.edu}.}
}
\theoremstyle{definition}

\newtheorem{remark}{Remark}

\newcommand{\bigCI}{\mathrel{\text{\scalebox{1.07}{$\perp\mkern-10mu\perp$}}}}

\theoremstyle{proposition}

\begin{document}

\conf{Workshop on Intelligent Aerial Robotics,\\ International Conference on Robotics and Automation, 2022}

\maketitle

\begin{abstract}
We propose a unified framework for coordinating Unmanned Aerial Vehicle (UAV) swarms operating under time-varying communication networks.
Our framework builds on the concept of graphical games, which we argue provides a compelling paradigm to subsume the interaction structures found in networked UAV swarms thanks to the shared local neighborhood properties.
We present a general-sum, factorizable payoff function for cooperative UAV swarms based on the aggregated local states and yield a Nash equilibrium for the stage games.
Further, we propose a decomposition-based approach to solve stage-graphical games in a scalable and decentralized fashion by approximating virtual, mean neighborhoods.
Finally, we discuss extending the proposed framework toward general-sum stochastic games by leveraging deep Q-learning and model-predictive control.
\end{abstract}

\section{Introduction and Related Work}

Unmanned Aerial Vehicle (UAV) swarms are currently emerging as a disruptive technology to reinforce many rapidly advancing research areas; mobile wireless networks \cite{skorobogatov2020multiple}, urban air mobility (UAM) systems \cite{thipphavong2018urban}, and intelligent transportation systems (ITS) \cite{saboor2021elevating} to name a few.
In practical deployments, many such applications demand the ability to coordinate UAV swarms of different scales spanning over large geographic regions with minimal communication infrastructure.
Despite the successful results shown by centralized UAV control in agile indoor swarms \cite{fernando2019formation}, such approaches can often come short attributing to the poor scalability and wireless signal degradation over large distances.
Although the decentralized swarm control frameworks increasingly emerge as an desirable alternative, the on-board computational and communication limitations can hinder global swarm state aggregation and optimization stages in control.
With the inherently non-stationary nature of real-world swarms stemming from the environment itself and the other robots, we stress that it is crucial to account for these limitations in one's control decision-making.

We propose a unified game-theoretic framework that subsumes the network and control level dynamics to drive a UAV swarm to a consensus on their actions by leveraging the available local information. 
Specifically, our framework builds on the concept of \textit{graphical games} to enable this.
Briefly, the graphical games reflect the notion that a multi-player game can be succinctly represented by a graph, and one's payoff only depends on a set of immediate neighbors \cite{kearns2013graphical}.
Further, we discuss the extensibility of our framework toward \textit{general-sum stochastic games} paradigm for controlling swarm robotics applications of different classes under time-varying communication networks.
The novelty of our research lies in its ability to directly synthesize the temporal communication topology of the robots into game-theoretical concepts for decision-making.
Furthermore, in contrast to more conventional distributed feedback control, game-theoretical formulations allows accounting for different levels of rationality in the robots \cite{Wen_Yang_Luo_Wang_2020}.
We believe that such a framework may highly appeal to applications where the UAVs are required to operate over large metropolitan areas while coordinating their actions through communication, i.e., synchronizing aerial vehicle take off and landing in UAM scenarios.
Our formulation further allows maximizing the individual's and the swarm's payoffs simultaneously through cooperation, which is of high interest for potential applications like autonomous urban mobility on demand (UMoD) and ITS.

Although extensive literature discusses swarm coordination through game-theoretic and graph-based approach\-es, the conjunction of the two is relatively sparse.
In \cite{olfati2007consensus}, Olfati-Saber et al. discussed reaching swarm consensus with dynamic interaction topologies through graph-based feedback control, albeit overlooking the game-theoretic aspects.
In contrast, our synthesis of game theory into the robots' communication topology allows using concepts from \textit{graphical models} to control swarms and analyze the consensus.
In \cite{Vamvoudakis_Lewis_Hudas_2012}, the authors presented the concept of \textit{differential graphical games} with \textit{continuous actions} for multi-agent reinforcement learning.
However, the direct application of the method on robot swarms is limited to quadratic reward functions.
In addition, \cite{ai2008optimality} and \cite{paraskevas2016distributed} proposed graphical game-theoretic methods for distributed mobile sensor coverage by conserving the energy; however, they overlook the communication uncertainties.

In contrast to classical methods, the mean-field type control approximates the global swarm state with a virtual \textit{mean} agent.
To eliminate the communication-intensive global state aggregation and forward prediction steps of mean-field approaches, Shiri et al. \cite{shi2021communicationaware} have proposed the use of federated learning for UAV swarm navigation.
Compared to our previous work in graphical-game theoretic coverage, we propose approximating the neighbors with higher-order connections with virtual \textit{mean} states, aggregated from local interactions.
\section{Approach}

\subsection{Switching Communication Network}
As the UAVs navigate over large distances, the wireless signal strength among them attenuates for multiple reasons; path loss, shadowing, and fading to name a few.
As a result, this leads to highly unpredictable topologies in the routing layer of the network.
\begin{figure}[t]
\centering
 \includegraphics[width = 0.42\textwidth, trim={0.3cm 0.2cm 0.3cm 0.2cm}, clip]{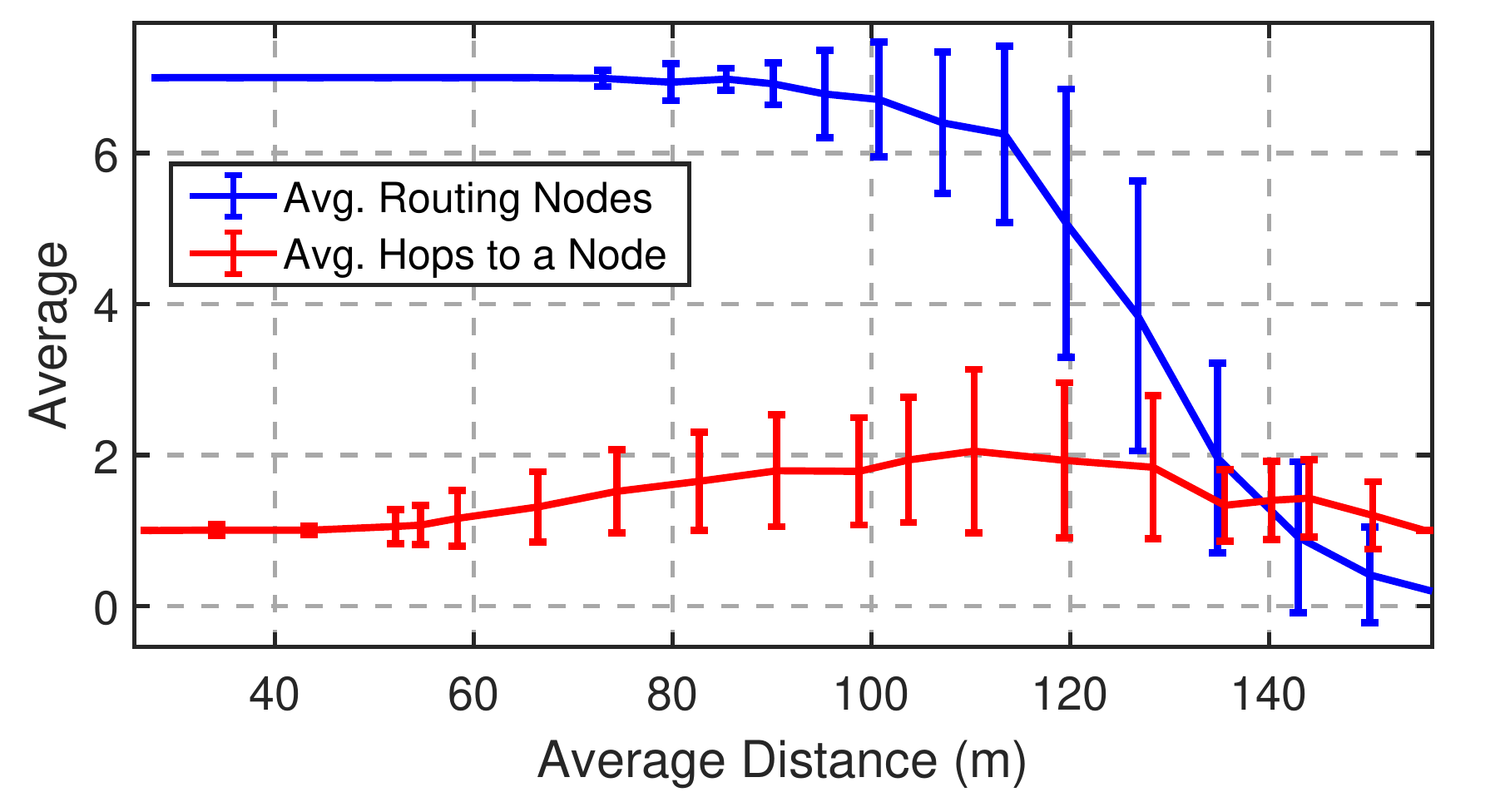}
\caption{The average number of directly communicable nodes (blue) and the average hop-count (red) against the physical distance. The $\inf$ hops have been ignored for the purpose of plotting. For the experiments we have used 7 UAV network nodes occupied with $16.02$ dBm power transmitters, and the \textit{Friis} propagation loss model under IEEE 802.11a protocol.}
 \label{fig:hops}
 \vspace{-10pt}
\end{figure}
This is apparent in our UAV ad-hoc wireless network simulation experiments results under the OLSR protocol presented in Fig. \ref{fig:hops}.
Briefly, OLSR is a proactive wireless routing protocol which periodically updates each node's routing table with a minimal bandwidth consumption, and thus, it has been often employed in UAV swarms and vehicular networks for practical deployments.
It can be seen that, as the UAVs move away from each others, the direct communication links disappear, and the hop-count between any two nodes to increase with a relatively high variance, especially at large distances.
This contradicts most of the fixed-radii-disk and nearest robot assumptions used in the literature to define the communicating neighborhoods.

Therefore, we stress that it is essential to take these real-world limitations into account when coordinating multi-robot systems in practice, particularly for communication intensive operations.
Motivated by this observation, we propose leveraging information from low-level network layer to compute time-varying dynamic neighborhoods for control.
More specifically, we define the neighborhood of any robot $i$ at any given time as $\mathcal{N}_i(h) = \{j \lvert \mathrm{Hops}(i,j) \leq h \}$, where $j \neq i$, $\mathrm{Hops}(i,j)$ is the number of communication hops to node $j$ according to $i$'s routing table.
Note that $h \geq 1$, and any robot belongs to its own neighborhood from a computer network perspective. 

\subsection{Cooperative Graphical Games for Robot Swarms}

The foundations of the graphical games were originally presented in \cite{kearns2013graphical}, and later extended with the theories from probabilistic inference in \cite{daskalakis2006computing, ortiz2020correlated}.
Although the notion of graphical games resembles that of collective dynamics-based approaches, only a few attempts have been made to employ the framework in swarm coordination, despite the success gained by the latter.

We argue that graphical games' neighborhood structures can seamlessly subsume the topologies of wireless communication networks to permit the robots to make decisions by naturally leveraging their local information flow.
In this work, we assume that all the robots in the swarm engage in a \textit{cooperative game} by sharing the same payoff function, and subsequently interested in maximizing some group objective.
This is synonymous with maximizing the profit of a UAV fleet in urban air mobility and the total signal strength in coverage control \cite{fernando2021coco} applications.
\subsubsection{Graphical game formulation}
First, we formulate the graphical game encountered in a single stage of a stochastic game involving a swarm of $N$ networked robots.
Let $\Gamma = \langle \mathcal{G}, \mathcal{S}, \mathcal{A}, \mathcal{M}\rangle$ define a stage graphical game, where $\mathcal{G} = \langle \mathcal{V}, \mathcal{E} \rangle$ is the undirected graph corresponding to the communication topology of the swarm, $\mathcal{S}$ is the global state space, $\mathcal{A}$ is the action space, and $\mathcal{M}$ is the set of payoff matrices associated with the robots. 
For the graph $\mathcal{G}$, we define the set of vertices $\mathcal{V}$, and edges $\mathcal{E}$ as the set of UAVs and their $k$-hop communication topology as acquired from the OLSR table.
More specifically, we let $(i,j) \in \mathcal{E}$ if, $\mathrm{Hops(i,j) \leq k}$ for all $i$.
We consider the global state space as $\mathcal{S} = \mathcal{S}_1 \times \mathcal{S}_2,  \times \dots  \times \mathcal{S}_N$, where $\mathcal{S}_i$ is the state space of the $i$-th UAV, $i \leq N$.
Similarly, the action space takes form $\mathcal{A} = \mathcal{A}_1 \times \mathcal{A}_2,  \times \dots \times \mathcal{A}_N$, where $\mathcal{A}_i$ is the actions available to the $i$-th UAV.
Additionally, we define aggregated neighborhood state and action spaces for any $i$ as $\mathcal{S}^\mathcal{N}_i = \prod_{j \in \mathcal{N}_i} \mathcal{S}_j$, and $\mathcal{A}^\mathcal{N}_i = \prod_{j \in \mathcal{N}_i} \mathcal{A}_j$, respectively.
Since any robot's payoff only depends on its neighborhood in graphical games, $\mathcal{M}_i$ in fact reduces to $\mathcal{M}_i: \mathcal{S}^\mathcal{N}_i \times \mathcal{A}^\mathcal{N}_i \xrightarrow{} \mathbb{R}$.

\subsubsection{Factorizing the payoff function}

We introduce a unary and a pairwise factorization of the payoff function $\mathcal{M}_i$ to accommodate 1). UAV application payoff functions with different localities, 2). computational feasibility and, 3) neighborhoods of varying sizes. 
Thus, we define the general-sum payoff function with strictly positive unary $\phi_u$ and pairwise $\phi_p$ factor potential functions,
\begin{equation}
    \mathcal{M}_i = \alpha_a \sum_{i \in \mathcal{N}_i} \phi_u(\mathcal{S}_i, \mathcal{A}_i) + \alpha_b \sum_{\substack{i,j \in \mathcal{N}_i\\j \neq i}} \phi_p(\mathcal{S}_i, \mathcal{A}_i, \mathcal{S}_j, \mathcal{A}_j),
    \label{eq:payoff}
\end{equation}
where $\alpha_a$ and $\alpha_b$ are two associated weights. 
The positiveness of the functions let us formulate the optimization as an energy minimization problem over an MRF \cite{fernando2021online}. 
This formulation of the payoff function is highly generalizable to a variety of applications: the unary functions can absorb contextual information that depends on the state of a single robot, and the pairwise functions can aggregate interdependent payoffs.
In addition, we have shown in \cite{fernando2021coco} this formulation can reach a mixed strategy NE in the graphical game by inferring the joint probability distribution over an MRF \cite{fernando2021coco}.

\subsection{Nash Equilibrium of the Graphical Game}
In this section, we discuss achieving the NE for the stage game $\Gamma$ through local game decomposition and probabilistic inference.
We highlight that this combination allows us solving the game in an efficient and a decentralized manner.
By definition, a NE of a graphical game results in the best response of any robot given the neighbors' actions, thus none of the robots has an incentive to unilaterally deviate \cite{kearns2013graphical}.
Let us first consider the Markov random field (MRF) defined on $\mathcal{G}$, and $\mathcal{V} = \{X_1,\dots, X_N\}$ where $X_i$ denotes the random variable (RV) associated with robot $i$.
Further, we define the scope of any RV as the corresponding robot's action space thus, $\mathrm{Scope}(X_\textbf{.}) = \mathcal{A}_\textbf{.}$.

\subsubsection{Local game decomposition}

To facilitate decentralized control, we decompose the game $\Gamma$ into $N$ independent local games $\{\Gamma_i | i\leq N\}$, each corresponding to a robot in the swarm. 
More specifically, we factorize the graph $G$ into $N$ subgraphs, in such a way that $\Gamma_i$'s graph structure is identical to the topology of $\mathcal{N}_i$.
Our solution stems from the idea that, we can solve each $\Gamma_i$ in an independent and a decentralized manner where each robot infers the best response actions for its neighborhood by leveraging the local information.

\begin{remark}
From the global Markovian property of an MRF, we can state that any two neighborhoods are conditionally independent given a separating subset of nodes, $\mathcal{N}_{\mathrm{sep}}$.
Thus, $\mathcal{N}_i \bigCI \mathcal{N}_j | \mathcal{N}_{\mathrm{sep}}$.
\label{rem:1}
\end{remark}
However, from remark \ref{rem:1}, it is clear that the independence between two local games holds upon fixing the latent value profile of $\mathcal{N}_{\mathrm{sep}}$. Consequently, if we can infer the same latent value profile for the separating RVs, the independence will be preserved.
From a game-theoretical perspective, this further ensures the local games reach the same NE, eliminating any ambiguities.
In practice however, for any robot $i$, the visibility limits to its neighborhood state space $\mathcal{S}^\mathcal{N}_i$. 
This restricts it from accurately inferring the actions of its neighbors $ X_j$, ($j \in \mathcal{N}_i, j \neq i$), who might rely on higher order neighbors' actions $X_k$, ($k \in \mathcal{V}/\mathcal{N}_i$); thus, leading each local game to reach different NE.
In \cite{vickrey2002multi}, the authors proposed an algorithm to solve the graphical game $\Gamma$ through decomposition, however we highlight that its direct synthesis would require extensive communication during the inference process to establish message passing among the robots.
In contrast, we propose an alternative, approximate method to reach the NE through local game decomposition for decentralized decision-making with a minimal communication overhead.

\subsubsection{Local game solution with virtual neighborhoods}
We focus on solving each $\Gamma_i$ separately, such that the inferred neighborhood action profile $\mathcal{A}_{i}^\mathcal{N}$ is a NE of the local game, and approximately preserves the game's independence.
We consider a special class of MRFs known as conditional random fields (CRF) which associate any RV $X_i$ with a corresponding \textit{observed} state variable; in this case $\mathcal{S}_i$.
For the clarity, let us reindex the RVs associated with the local game $\Gamma_i$ as $X^i_1, \dots, X^i_j, X^i_{|\mathcal{N}_i|}$.
Given the states of the neighborhood $\mathcal{S}^\mathcal{N}_i$, and the latent space $\mathcal{A}^\mathcal{N}_i$, our objective is to find the joint probability distribution over the RVs $P(X^i) = P(X^i_1, \dots, X^i_j, \dots, X^i_{|\mathcal{N}_i|} | \mathcal{S}^\mathcal{N}_i)$.
Further, for a given state space $\mathcal{S}_i^{\mathcal{N}_i}$, payoff function becomes $\mathcal{M}_i(X^i)$.
\begin{algorithm}
\SetAlgoLined
 Populate $\mathcal{N}_i$ using the routing table for $i$\\
 For all $ j \in \mathcal{N}_i$, $j \neq i$ Aggregate $\mathcal{N}_j$ \\
 For all $j \neq i$, $\mathcal{N}_j - \mathcal{N}_i \neq \emptyset$ approximate $\bar{\mathcal{S}}_j \in \bar{\mathcal{S}}^\mathcal{N}_i$ \eqref{eq:approx} \\
\For{$j \leftarrow 1 \dots |\mathcal{N}_i|$} {
     $Q_j(X_j) \leftarrow \frac{1}{Z_j} \exp \big\{ - \alpha \sum_{j}\phi_u(x_j, \bar{\mathcal{S}}^\mathcal{N}_i)$ \big\}
     }
 \While{KL$\big[Q_{old}(X^i) | Q(X^i)\big] \geq \delta$}{
    $Q_{old}(X^i) = Q(X^i)$ \\
    Choose $X^i_j$ from $\{X^i_1, \dots , X^i_{|\mathcal{N}_j|}\}$ \\
    \For{$x_j \in \mathcal{A}_j$ } {
    $Q_j(x_j) \leftarrow \frac{1}{Z_j} \exp \{-M_i \big(x_j, X^i_k, \bar{\mathcal{S}}^\mathcal{N}_i \big) \} $ \\
    }
    $Q(X^i) \leftarrow \prod_{j}Q_j(X^i_j)$
 }
 \caption{Solving Local Game $\Gamma_i$}
 \label{algo}
\end{algorithm}
Following the formal definition of MRF, we define the joint probability distribution over the RVs in $\Gamma_i$ as
\begin{equation}
    P(X^i | \mathcal{S}_i^{\mathcal{N}_i}) = P(X^i) = \frac{1}{Z}\exp{\Big[-M_i \big(X^i_j, X^i_k \big) \Big]},
\end{equation}
where $Z$ is a partition function to normalize the distribution.

To reduce the conditional dependence of any node $j \in \mathcal{N}_i$, $j \neq i$ on higher order neighbors, we approximate it with a representative node by averaging its neighborhood state space.
Specifically, we approximate all $j \in \mathcal{N}_i$ where $\mathcal{N}_j - \mathcal{N}_i \neq \emptyset $ by a \textit{virtual} \textit{mean} robot with state $\bar{\mathcal{S}}_j$ where,
\begin{equation}
    \bar{\mathcal{S}}_j = \frac{1}{|\mathcal{N}_j|} \sum_{k \in \mathcal{N}_j} S_k.
    \label{eq:approx}
\end{equation}


We propose computing the joint probability distribution over the CRF using \textit{mean-field approximation} \cite{koller2009probabilistic} where the joint distribution is assumed to be a product of the marginals. Thus,
\begin{equation}
    P(X^i) \approx Q(X^i) = \prod_{j \in \mathcal{N}_i} Q_j(X^i_j).
\end{equation}
where, $\bar{\mathcal{S}}_i^{\mathcal{N}_i}$ is the approximated neighborhood state space.
The \textbf{Algorithm} \ref{algo} summarizes the steps of our approach.
We initialize the marginal distributions using unary potential functions in line 5.
The convergence of the algorithm is measured by the Kullback-Leiber (KL) divergence of the approximating distribution $Q(X^i)$ between two adjacent iterations, and terminate as it reaches a small predefined threshold $\delta$, ($0 \leq \delta << 1$). 



\section{Experiments and Results}
We evaluated the proposed approach on a communication-aware coverage scenario, where a swarm of UAVs are deployed to establish an ad-hoc mobile network over a large geographic region of interest (ROI).
We define the ROI as the concentration ellipsoid $\mathcal{R}$ of a predefined bi-variate Gaussian distribution, and $p(r)$ as the probability density of any point $r \in \mathcal{R}$.
The unary term of \eqref{eq:payoff} integrates the \textit{maximum} \textit{expected} signal strength imparted by the neighborhood over the ROI, and the pairwise term computes the \textit{expected} signal strength between any two neighbors.
Therefore,
\begin{equation}
    \phi_{u} = \int_{r} \max \big\{ \psi_{R}(\mathcal{S}_i, r), \psi_{R}(\bar{\mathcal{S}}_{j}, r)  \big\} p(r), \forall j \neq i,
    \label{eq:unary}
\end{equation}
\begin{equation}
    \phi_{p} = \psi_{R}(\mathcal{S}_i, \bar{\mathcal{S}}_j), \forall i,j \in \mathcal{N}_i, i\neq j,
    \label{eq:pairwise}
\end{equation}
where, $\psi_{R}$ is the expected received signal strength calculated using the \textit{Friis} propagation loss model with transmission power $T_0$, reference loss $L_0$ and distance between the any two points $d$ such that $\psi_{R} = T_0 - \big\{ L_0 + 10n . \log(d) \big\}$.
Throughout the experiments, we fixed the maximum allowed communication hops $h$ to 2, tuning parameters $\alpha_a=1$ and $\alpha_b=0.001$.
\begin{figure}[t]
\centering
\subfigure[]
 {\includegraphics[width = 0.235\textwidth, trim={0.3cm 0cm 0.5cm 0.2cm}, clip]{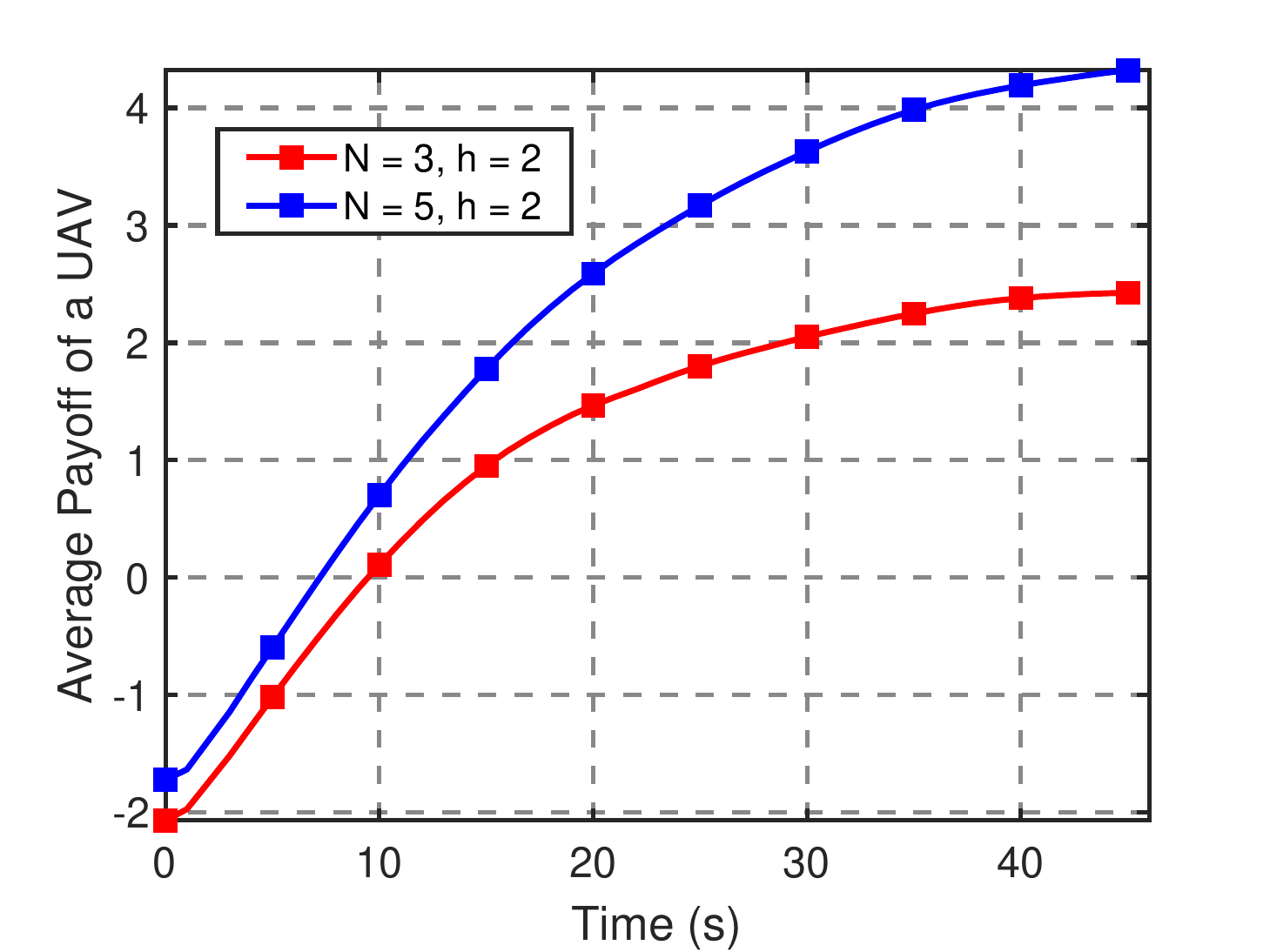}}
\subfigure[]{
 \includegraphics[width = 0.235\textwidth, trim={0.3cm 0cm 0.5cm 0cm}, clip]{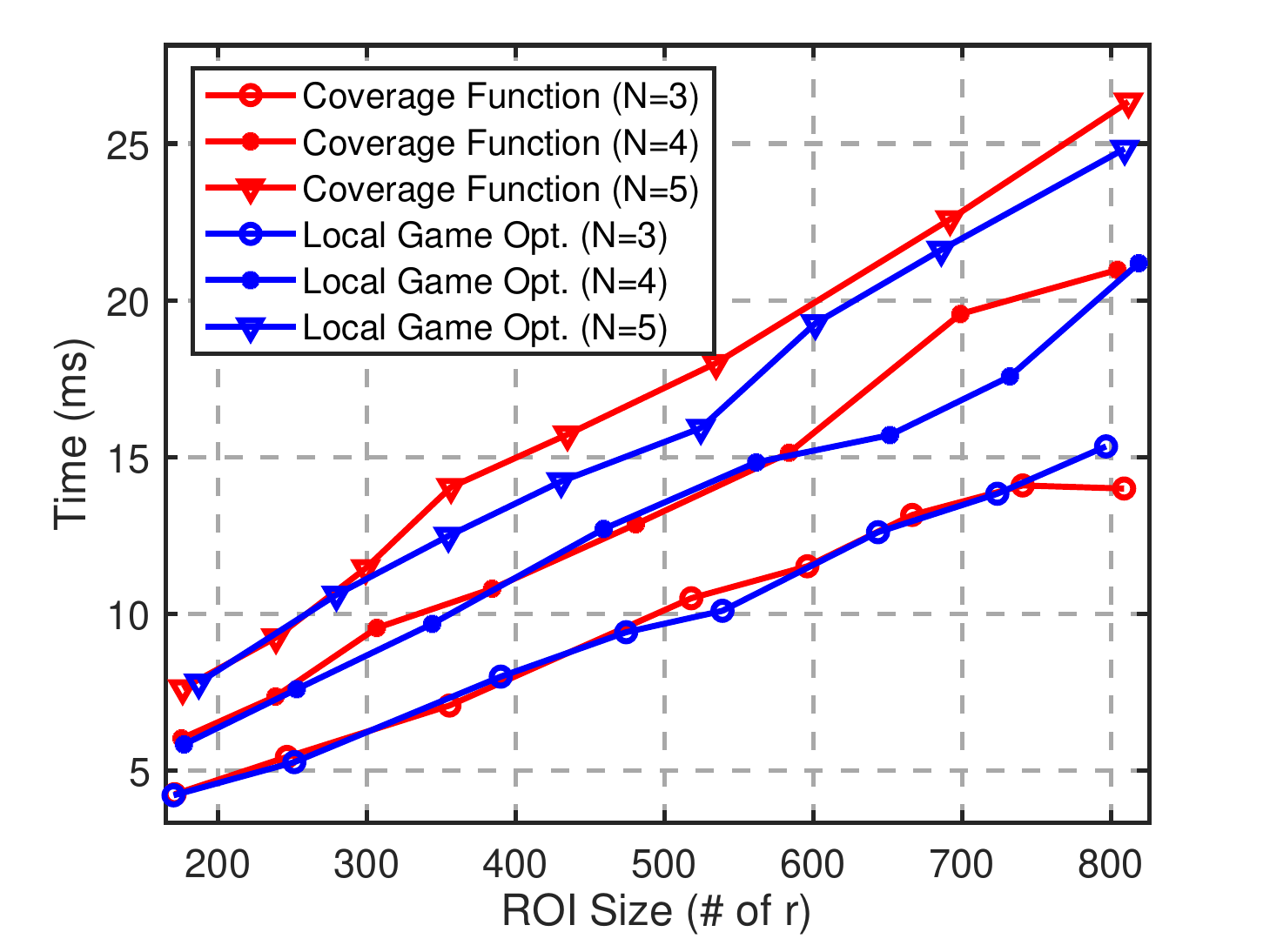}}
\caption{(a). Change of average payoff of a robot against the time with iterative local game optimization in real-time. (b). Computational time for local game optimization (blue) and payoff function integration (red).}
 \label{fig:results}
 \vspace{-10pt}
\end{figure}
We use the ROSNS3 simulator from \cite{fernando2021coco} for the experiments, however, the computational power required for real-time network-physics calculations limits us to small size swarms \footnote{ROSNS3 Simulator:\href{https://github.com/malintha/rosns3} {https://github.com/malintha/rosns3}}.
In Fig. \ref{fig:results}(a) we show the average payoff of a UAV operating over the ROI against time and the number of robots in the swarm.
It can be seen that the proposing method can successfully drive the robots to consensus while increasing the individual and the swarm payoff.
Although, we were not able to observe a significant difference in the payoff for the virtual neighbor states compared to using the unchanged states for swarms up to 5, we believe that this will become more apparent in large-scale swarm simulations. 
Fig. \ref{fig:results}(b) shows the computational time for computing the payoff function and the optimization stages. 
We can observe that, despite the computational complexity of the payoff function, which is highly dependent on the size of the ROI, the local game optimization to converge within a few milliseconds.

\section{Future Work}
In our ongoing research, we query the presented framework's extensibility toward coordinating swarms through the paradigm of \textit{stochastic graphical games} or, in broader terms, multi-agent reinforcement learning (MARL).
More accurately, we propose two lines of research for that purpose: 1). combination of the local game's solution with model predictive control (MPC) and 2). use of a deep neural network to approximate the Nash Q-function of the game.
The former approach couples the stage game solution with dynamic programming and, thus, heavily depends on the computational feasibility of the current approach.
Thanks to the rapidly converging nature of our local game optimization algorithm for finite size neighborhoods as shown, it highly appeals to the MPC paradigm.
Additionally, by learning an aggregation graph-neural network to approximate \eqref{eq:approx}, and to predict the future payoffs based on connectivity, our approach can further benefit controlling communication-intensive UAV swarm applications.
This method further alleviates performing heavy computations of the payoff function using non-linear function approximaters. 

The second approach combines the stage game optimization with deep Q-learning to train a robust policy from the observations instead of choosing actions directly.
We are motivated by the promise of Nash Q-learning for \textit{two player general-sum games} \cite{hu2003nash, yang2018mean}, and believe our graphical game-theoretic framework with virtual neighborhoods and factorized payoff functions has the potential to extend the itself toward MARL paradigm in a scalable and decentralized fashion.
We further highlight that these two directions can further rule out the requirement of accurate states of the robots are being directly  observable.

\bibliographystyle{ieeetr}
\bibliography{root}

\end{document}